\pdfoutput=1
\documentclass[11pt]{article}

\usepackage{EMNLP2022}

\usepackage{times}
\usepackage{latexsym}
\usepackage{hyperref}
\usepackage{url}
\usepackage{textcomp}
\usepackage{natbib}
\usepackage{graphicx}
\usepackage{tabularx}
\usepackage{appendix}
\usepackage{stfloats}
\usepackage{floatrow}

\definecolor{ngreen} {RGB}{98,158,31}
\definecolor{nred}   {RGB}{224,0,0}

\usepackage{subfig}
\usepackage{amsmath}
\newcommand{\argmax}[1]{\underset{#1}{\operatorname{arg}\,\operatorname{max}}\;}

\usepackage{float}
\restylefloat{table}
\usepackage{makecell}
\graphicspath{{images/}}
\usepackage{multirow}
\usepackage{tablefootnote}
\usepackage{longtable}

\makeatletter
\def\maketag@@@#1{\hbox{\m@th\normalfont\normalsize#1}}
\makeatother

\usepackage[T1]{fontenc}
\usepackage[utf8]{inputenc}

\usepackage{microtype}

\usepackage{inconsolata}

\title{Generating image captions with external encyclopedic knowledge}

\author{Sofia Nikiforova, Tejaswini Deoskar, Denis Paperno, Yoad Winter \\
Utrecht University, Utrecht, the Netherlands \\
\texttt{\{s.nikiforova, t.deoskar, d.paperno, y.winter\}@uu.nl}
}

\begin{document}
\makeatletter
  \@namedef{figure}{\killfloatstyle\def\@captype{figure}\FR@redefs
    \flrow@setlist{{figure}}%
    \columnwidth\columnwidth\edef\FBB@wd{\the\columnwidth}%
    \FRifFBOX\@@setframe\relax\@@FStrue\@float{figure}}%
\makeatother

\maketitle

\begin{abstract}
Accurately reporting what objects are depicted in an image is largely a solved problem in automatic caption generation. The next big challenge on the way to truly humanlike captioning is being able to incorporate the context of the image and related real world knowledge. We tackle this challenge by creating an end-to-end caption generation system that makes extensive use of image-specific encyclopedic data. Our approach includes a novel way of using image location to identify relevant open-domain facts in an external knowledge base, with their subsequent integration into the captioning pipeline at both the encoding and decoding stages. Our system is trained and tested on a new dataset with naturally produced knowledge-rich captions, and achieves significant improvements over multiple baselines. We empirically demonstrate that our approach is effective for generating contextualized captions with encyclopedic knowledge that is both factually accurate and relevant to the image.
\end{abstract}

\section{Introduction}
\label{sec:intro}

\begin{figure}[ht]
    \begin{center}
    {\includegraphics[scale = 0.49]{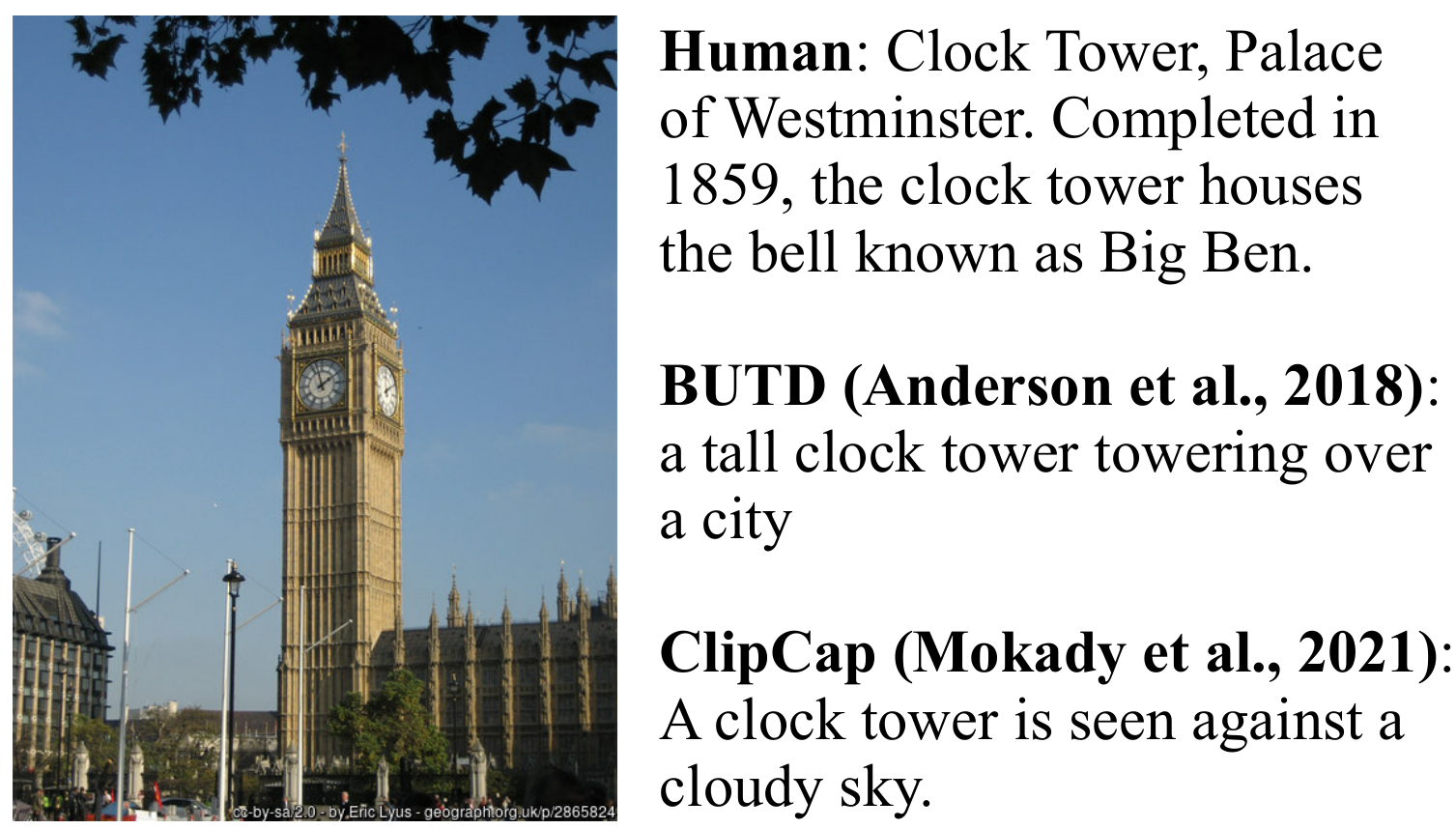}}
    \caption{Human-generated caption vs. captions generated by standard automatic captioning systems.} 
    \label{fig:big_ben}
    \end{center}
\end{figure}

Most modern image captioning systems are designed to produce a straightforward description of the visual content of the image. They abstract away from the image context and have no access to information that is not directly inferred from the image. By contrast, humans can extend beyond a purely visual description and produce a caption that is influenced by the image context and informed by real world knowledge. As a result, human-generated and automatically generated captions for the same image can be drastically different, as seen in Figure~\ref{fig:big_ben}. This contrast motivates the task of \emph{contextualized} image captioning, which sets the goal of automatically generating captions that would account for relevant image-external knowledge.

In this paper we propose an approach to contextualizing an image captioning system and enriching it with a wide range of encyclopedic knowledge. Our key contribution is in constructing an image-specific knowledge context that consists of relevant facts from an open-domain knowledge base, retrieved based on image location metadata. We bridge the gap between two lines of research in contextualized image captioning: one utilizes knowledge from external data sources, establishing its relevance through image recognition \citep{bai2021explain,huang2020boost}, the other uses geographic metadata to identify the names of geographic entities in and around the image \citep{nikiforova-etal-2020-geo}. To the best of our knowledge, we are the first to combine these two approaches --- we use geographic metadata to access an external knowledge base and obtain various kinds of non-geographic encyclopedic facts relevant to the image, with their subsequent integration into the captioning pipeline.

We implement the proposed approach in a knowledge-aware caption generation system with image-external knowledge incorporated at both the encoding and decoding stages.
We also present a new dataset of images with naturally created knowledge-rich captions and corresponding geographic metadata. Our experiments on this dataset show the effectiveness of our approach: the system greatly outperforms multiple baselines in commonly used captioning metrics and, crucially, the correctness of the generated facts.

\section{Related Work}
\label{sec:background}

Two main aspects that define each approach to contextualized image captioning are (i) the source of external knowledge and the way to identify the data relevant to a particular image, and (ii) the method of incorporating external knowledge into caption generation.

\paragraph{External knowledge source} In a popular subtask of news image captioning, captions are generated for images that accompany news articles \citep{zhao2021boosting, hu2020icecap, Tran2020, Jing2020, Chen2020, Biten2019}. Naturally, the article texts themselves are the main source of context for captioning, supplying information about important events and entities. In a more general case images are not paired with the relevant context directly. A common way to connect images to an external source of knowledge is to use an object detection mechanism to identify objects in the image and then use their labels to query a database. In \citet{huang2020boost, Zhou2019, wu2017image} detected labels are used to extract useful information about common objects featured in the images (such as ``dog'', ``pot'', ``surfboard'') from ConceptNet \citep{Speer2017} and DBpedia \citep{Auer2007}. \citet{Zhao2019} use Google Cloud Vision APIs to identify not only common objects but also entities (people, car brands, etc). In \citet{bai2021explain} custom classifiers are trained to detect specific image attributes (e.g. the author, the artistic style), which are then used to retrieve relevant information from Wikipedia. These approaches exclusively use the visual content of the images to contextualize the captioning process. Thus, the extent of contextualization is limited by the quality of the object detection algorithms, and the potential benefit of utilizing additional data (e.g. image metadata) is left unexplored.

Certain types of image metadata can be used to build upon general object detection and identify specific entities and events in the image. \citet{Lu2018} use the time metadata of the image (the date when a given photo was taken) and its associated tags to collect similar photographs and to retrieve the names of relevant entities (e.g. people depicted in the image) from their captions. In \citet{nikiforova-etal-2020-geo}, geographic metadata (latitude and longitude coordinates of the image location) is used to extract information about the surrounding objects from a geographic database, which allows their system to refer to concrete locations relevant to the image in the generated captions. These papers demonstrate the effectiveness of using image metadata for contextualized captioning but are limited to establishing the names of relevant entities, and do not utilize them to get further data that could be useful for generating even more informative captions.

In contrast to the works described above, we use image metadata as a grounding ``anchor'', with which we can not only identify entities relevant to the image, but also retrieve a wide range of related encyclopedic knowledge from an external database.
Specifically, we use geographic metadata, which has the benefit of being easily available for many real-life photographs due to the built-in GPS in modern cameras and phones, making it easier to collect the data for training and testing the system.

\paragraph{Incorporating external knowledge into caption generation} There are two dominant methods of incorporating external knowledge into the caption generation process: template-based and context-based. In template-based approaches, a caption is generated with placeholder token slots that are later filled with the most fitting named entities extracted from an external knowledge source \citep{bai2021explain, Jing2020, hu2020icecap, Biten2019}. This is an especially common technique in news image captioning, where named entities are taken from the news article associated with the image. Still, the straightforward fill-in-the-slot method can be problematic if none of the available entities fit the already generated placeholder slot.

In context-based approaches, external knowledge informs the caption generation process along with the image features, influencing the choice of produced tokens. For example, \citet{Zhou2019} extract ConceptNet terms related to the image and use their embeddings to initialize the caption generation module. \citet{huang2020boost} also use ConceptNet to identify relevant external knowledge and increase the output probabilities of the vocabulary tokens if they match the extracted entities. The downside of context-based models is their inability to generate tokens that are present in the external knowledge but happen to be out-of-vocabulary for the generator's language model, which is common for named entities.

Our model, like several other approaches \citep{Chen2020, Tran2020, nikiforova-etal-2020-geo, Whitehead2018}, combines characteristics of both template-based and context-based methods. Similarly to template-based architectures, some of the tokens produced by our caption generation model are taken directly from external knowledge sources, and as in the context-based methods, external knowledge in our model influences the generation of regular vocabulary words.

\section{Image caption dataset}
\label{sec:dataset}

For the purposes of training a knowledge-aware image captioning system, we require a dataset of images with knowledge-rich captions. Standard image captioning datasets, such as MSCOCO \citep{Lin2014} or Flickr30k \citep{Young2014}, contain heavily curated captions, purposefully stripped of any contextualization and references to encyclopedic knowledge. Moreover, they lack image metadata, which is critical for our approach. Therefore, we compile our own dataset of 7128 images with the associated naturally created captions and image location metadata. The source of our data is the website of the Geograph project\footnote{ \url{http://www.geograph.org.uk/}}, which contains a large number of photographs with English-language captions and rich metadata, including the geographic coordinates of the photograph locations. The images we selected for our dataset depict geographic entities, mostly historical man-made structures and buildings, to ensure that each caption contains at least one encyclopedic fact, for example, ``Theatre Royal Haymarket. Dating back to 1720''. More details of the dataset, including its split into train, validation and test sets, are given in Appendix~\ref{sec:appendix_split}.

\section{Approach}
\label{sec:approach}

Our approach begins with identifying real world entities that are related to the image. Specifically, we use the coordinates of where the photograph was taken to retrieve a set of geographic entities around that location (Section~\ref{sec:geocontext}). Then, we collect a wide range of relevant encyclopedic facts. We extract them from an open-domain knowledge base in the form of $<$subject, predicate, object$>$ triples, where subjects are the previously identified geographic entities (Section~\ref{sec:knowledge_context}). Finally, we integrate the collected knowledge into an otherwise standard encoder-decoder image captioning pipeline to produce knowledge-aware captions (Section~\ref{sec:model_architecture}).

\subsection{Geographic Context}
\label{sec:geocontext}

We adapt an approach from \citet{nikiforova-etal-2020-geo} to retrieve geographic entities relevant to the image from the OpenStreetMap\footnote{\url{https://www.openstreetmap.org/}} database and thus to construct an image-specific geographic context.
Formally, the geographic context $G$ of a given image is a set of $n$ geographic entities ($e_1 \dots e_n$) located within a radius $r$ from the image location. Based on preliminary experiments, we set $n$ at 300 and $r$ at 1 kilometer as the hyperparameters of our system.

\begin{figure}[H]
    \begin{center}
    \includegraphics[scale = 0.58]{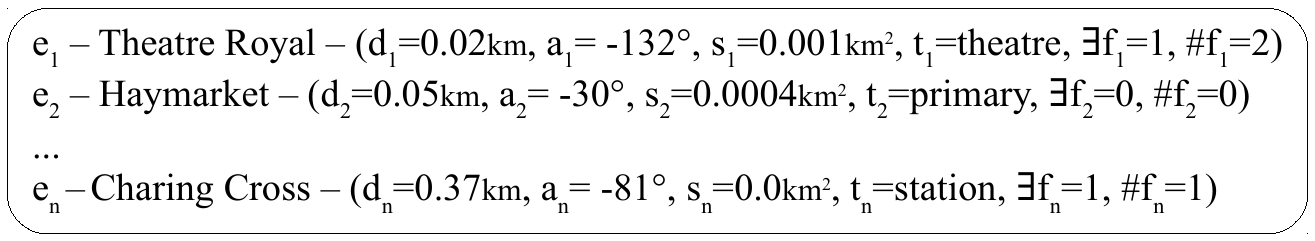}
    \caption{A fragment of a geographic context.}
    \label{fig:geo_context}
    \end{center}
\end{figure}

Each geographic entity $e_i$ is associated with its name and a set of geographic features: distance $d_i$ and azimuth $a_i$ between the entity and the image location, the entity's size $s_i$ and type $t_i$. In addition to that, we introduce two new features, intended to reflect the salience of the entity through the amount of information available about it in a knowledge base: a binary indicator $\exists f_i$ that shows whether or not the entity corresponds to any facts in the knowledge context, and the number of facts $\#f_i$ that correspond to the entity in the knowledge context. A sample fragment of a geographic context, with the entities mapped to their names and features, is shown in Figure~\ref{fig:geo_context}.
The features are combined in vector representations for the entities, called ``geographic embeddings''. For an entity $e_i$ a geographic embedding is computed as follows:
\begin{equation}\label{eq:geoemb}
\begin{split}
    \textsc{GeoEmb}(e_i) = \text{ Concat}[d_i,\:norm(a_i),\:s_i,\\
    \exists f_i,\:\#f_i, Emb_t(t_i)]
\end{split}
\end{equation}

where $norm$ is an azimuth normalization function, $Emb_t$ is an embedding function for the entities' types. For implementation details, see Appendix~\ref{sec:appendix_implementation}. Our experiments show that our straightforward concatenation strategy achieves the same performance level as the more complicated approach employed by \citet{nikiforova-etal-2020-geo}.

\subsection{Knowledge Context}
\label{sec:knowledge_context}

For a given image with the geographic context $G$, the knowledge context $K$ is defined as a set of $m$ facts ($f_1 \dots f_m$) about the entities in $G$.

We obtain the facts from the DBpedia knowledge base, where they are stored as triples of the form $<$subject, predicate, object$>$. First, we select all the facts, in which the subject is one of the geographic entities from $G$, e.g. $<$Theatre Royal, built\_in, 1720$>$, $<$Theatre Royal, architect, John Nash$>$, $<$Theatre Royal, rebuilt, 1879$>$, etc. Then, we train a logistic regression model to rank the facts based on how likely they are to be mentioned in the caption. The model takes into account the fact's predicate, the ranking of the fact's subject in the geographic context and its geographic features. The top $m$ facts of the ranked list constitute the knowledge context of the image, with $m$ as another hyperparameter of the system, which we set at 50.

Knowledge contexts of all the images in our dataset contain altogether facts with 1542 unique predicates, 336 of which appear at least once in the ground truth captions. The detailed statistics of the collected knowledge data (e.g. fact and predicate distribution per entity, image and caption) are provided in Appendix~\ref{sec:appendix_knowledge_stats}.
\begin{figure}[ht]
    \begin{center}
    \includegraphics[scale = 0.8]{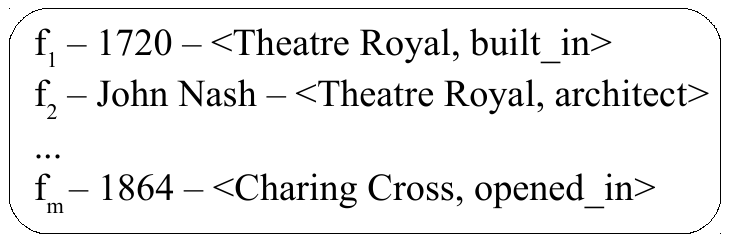}
    \caption{A fragment of a knowledge context.}
    \label{fig:know_context}
    \end{center}
\end{figure}

Figure~\ref{fig:know_context} shows a fragment of the knowledge context corresponding to the geographic context in Figure~\ref{fig:geo_context}.
As seen in the figure, we modify the way the original DBpedia facts are represented. Instead of the $<$subject, predicate, object$>$ triples, we treat the objects as labels of the facts (similarly to the names, such as ``Theatre Royal'', that act as labels of the geographic entities). In our approach, it is the fact objects that constitute the `vocabulary' of fact-related tokens that appear in captions. For example, in the caption ``Theatre Royal Haymarket. Dating back to 1720'', the fact $<$Theatre Royal, built\_in, 1720$>$ is realized through the presence of the fact's object ``1720'', while ``Theatre Royal'' comes from the geographic context and ``dating back'' comes from the regular vocabulary. So, given the contexts in Figures~\ref{fig:geo_context} and \ref{fig:know_context}, this caption will actually be processed by the system as ``$e_1$ $e_2$. Dating back to $f_1$''.

Thus, each fact in the knowledge context is represented through its object as a label, which can appear in a caption, and through the $<$subject, predicate$>$ pair, which reflects the fact's meaning. We encode this meaning in a vector form, a ``fact embedding'', which is computed as follows:
\begin{equation}\label{eq:factemb}
    \textsc{FactEmb}(f_i) = \textsc{GeoEmb}(e_i) + Emb_p(p_i)
\end{equation}
where $e_i$ is the subject of the fact $f_i$ (an entity from the geographic context), $p_i$ is its predicate and $Emb_p$ is an embedding function for the predicates. Predicate embeddings are initialized randomly and updated in an end-to-end fashion during the training of the captioning model. 

This approach to encoding facts in the knowledge context provides the captioning system with the information it needs to select an appropriate fact based on the previously generated tokens. For any given fact, the system can take into account whether or not its subject is already present in the caption and estimate whether the previous caption tokens are consistent with the fact's predicate. One possible alternative for encoding structured external knowledge is applying graph embedding methods (such as a popular TransE algorithm \citep{bordes2013translating}) to a knowledge graph, and then utilizing the embeddings of the graph nodes (entities) and edges (relations) during text generation. Our approach is less time- and resource-consuming, since it requires no extra pre-training: predicate embeddings are trained together with the captioning model, we re-use geographic embeddings to encode fact subjects, and embeddings of the fact objects are a linear combination of the predicate and subject embeddings. Moreover, using geographic embeddings for encoding the fact subjects provides additional grounding to the knowledge context and a direct link between the knowledge context and the geographic context.

\subsection{Model Architecture}
\label{sec:model_architecture}

\begin{figure*}[ht]
    \centering
    \begin{center}
    \includegraphics[scale = 0.56]{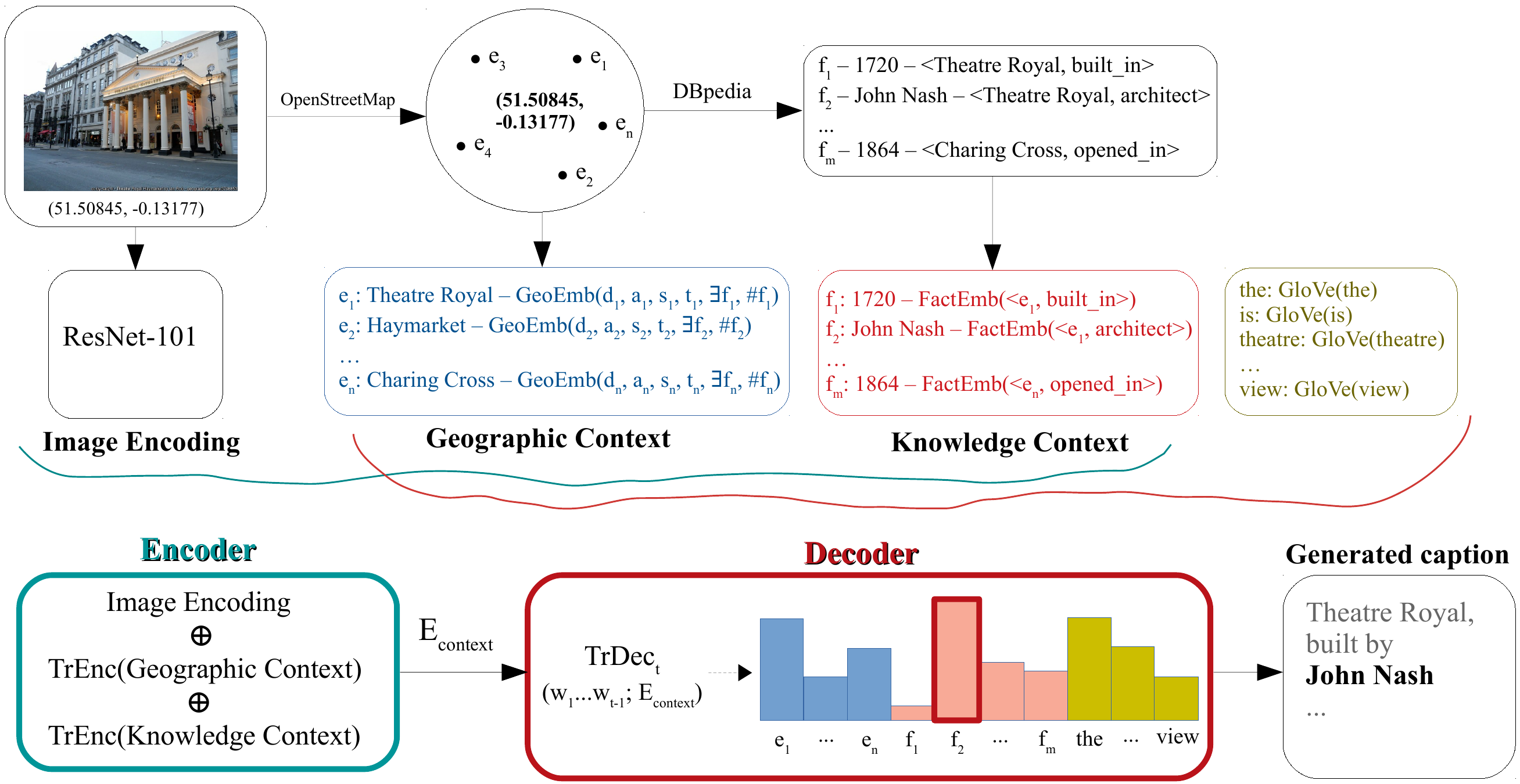}
    \caption{An overview of the system architecture (best viewed in color).}
    \label{fig:diagram}
    \end{center}
\end{figure*}

Our caption generation system is an end-to-end trainable neural network with an encoder-decoder architecture. The overview of the system's architecture is given in Figure~\ref{fig:diagram}. As shown in the figure, both encoder and decoder make use of the external knowledge to produce knowledge-rich captions. See Appendix~\ref{sec:appendix_implementation} for concrete implementation details.

\paragraph{Encoder}
The encoder's function is to convert input data into an informative representation that is subsequently used by the decoder to generate a caption. In a standard image captioning pipeline, the input data consists only of the image itself, and its encoding is a dense representation of its visual features. In our system, we also use geographic and knowledge contexts as the additional sources of input data.

For the image encoding $E_{img}$, we use a deep convolutional neural network (CNN), pre-trained on an image classification task. The particular CNN that we use is ResNet-101 \citep{He2016}, trained on the images from the ImageNet database \citep{Russakovsky2015}.
In addition, we encode information contained in the geographic and knowledge contexts. First, each of their elements is embedded with the embedding functions introduced in Sections~\ref{sec:geocontext}, \ref{sec:knowledge_context}:
\begin{equation}\label{eq:embed_contexts}
\begin{split}
    EmbG = (&\textsc{GeoEmb}(e_1)\dots\\
    &\dots\textsc{GeoEmb}(e_n)), e_i \in G \\
    EmbK = (&\textsc{FactEmb}(f_1)\dots \\
    &\dots\textsc{FactEmb}(f_m)), f_i \in K \\
\end{split}
\end{equation}
They are subsequently encoded with two separate Transformer encoders (TrEnc), with a standard structure proposed in \citet{10.5555/3295222.3295349}. 
Finally, we concatenate the encodings of the image, the geographic context and the knowledge context:
\begin{equation}\label{eq:combine_encodings}
\begin{split}
    E_{context} &= \\
    = \text{Concat}&[\text{\small{$E_{img}$, \textsc{ TrEnc}(EmbG), \textsc{ TrEnc}(EmbK)}}]
\end{split}
\end{equation}
The result of the concatenation is the combined representation of the visual features of the image and the relevant information from the geographic and knowledge contexts.

\paragraph{Decoder}
The decoder accepts the combined context representation from the encoder and generates an output sequence --- the caption. The goal of the decoder is to produce a caption that would be fitting to the image and include accurate references to the external knowledge. 
The decoder generates a caption token by token, at each step $t$ taking into account the previously generated tokens $w_1\dots w_{t-1}$ and the context representation $E_{context}$. In the process, each input token is represented by a sum of its vector embedding and the encoding of its position in the sequence.
\begin{equation}\label{eq:pos_emb}
    PosEmb(w_i) = Emb(w_i) + Pos(w_i)
\end{equation}
We distinguish three types of tokens: regular vocabulary words, names of geographic entities (the names mapped to entities in the geographic context) and fact-related tokens (the objects of the facts in the knowledge context). We use pre-trained GloVe word embeddings \citep{Pennington2014} for the regular vocabulary tokens. However, conventional pre-trained embeddings are not particularly informative for the geographic entity names and the fact-related tokens, as their semantics is conveyed poorly by their distribution patterns (e.g. the various contexts in which the token ``1720'' appears in a large scale corpus are too diverse for a precise representation of ``1720'' as the year when Theatre Royal was built). It is important that the decoder can utilize information about their most meaningful characteristics: physical properties of the geographic entities and the facts' subjects and predicates. For this reason, we use the \textsc{GeoEmb} and \textsc{FactEmb} embedding functions to represent them.
\begin{equation}\label{eq:token_embed}
 Emb(w_i) =
\begin{cases}
   \textsc{GeoEmb}(w_i), \text{if } w_i \in G \\
   \textsc{FactEmb}(w_i), \text{if } w_i \in K \\
   \textsc{GloVe}(w_i), \text{otherwise} \\
\end{cases}
\end{equation}
We employ a Transformer decoder (TrDec) with a standard structure. It attends to the positional embeddings of the previously generated tokens and to the encoder's output, the combined representation of the image contexts.
\begin{equation}\label{eq:decoder_out}
\begin{split}
    h_t = \textsc{TrDec}(PosEmb(w_{1...t-1}); E_{context})
\end{split}
\end{equation}
In a standard captioning pipeline, the output of the decoder $h_t$ is then passed to a final linear layer that acts as a classifier, estimating the probability distribution over all the tokens in the vocabulary $V$. The vocabulary is usually fixed and consists of the words or subwords from the training dataset. In our case, captions also include entity names and facts from the geographic and knowledge contexts, which are image-specific, and therefore, not all relevant entity names and fact-related tokens would necessarily be a part of $V$. So, we modify the last stage of the decoding process by computing three sets of scores: the scores for the vocabulary tokens $v_1...v_k \in V$, the scores for the geographic entity names $e_1...e_n \in G$ and the scores for the facts $f_1...f_m \in K$.
\begin{equation}\label{eq:scores}
    \begin{split}
    y_{v_1}...y_{v_k} &= (h_t \circ p_{ind}\:W_{pred}) \; W_{vocab} \\
    y_{e_1}...y_{e_n} &= (EmbG \; \textsc{diag}(h_t))\: \vec{w}_{geo} \\
    y_{f_1}...y_{f_m} &= ((EmbK \circ g_{ind}) \; \textsc{diag}(h_t))\: \vec{w}_{f} \\
    \end{split}
\end{equation}
where $W_{vocab}$ and $W_{pred}$ are trainable linear transformation matrices, $\vec{w}_{geo}$ and $\vec{w}_{f}$ are trainable linear transformation vectors, $\textsc{diag}(h_t)$ denotes a diagonal matrix with the $h_t$ vector in the main diagonal, $\circ$ stands for a Hadamard product. 

The predicate indicator $p_{ind}$ and the geo entity indicator $g_{ind}$ provide additional contextualization to the generation process\footnote{We have also experimented with simpler architectures, without the predicate and geo entity indicators. The performance of these alternative systems was overall lower, especially in the accuracy of the generated encyclopedic facts. See Appendix~\ref{sec:appendix_alt} for more details.}. The geo entity indicator specifies for each knowledge context fact whether its subject has already been mentioned in the caption at a given time step (1 for `has been mentioned', 0 for `has not been mentioned'). Facts with the subjects that are already in the caption would generally get a higher generation probability assigned to them by the model. 

The predicate indicator ensures that the external knowledge has influence over the generation of regular vocabulary words. It reflects for each of the known predicates whether the knowledge context contains a fact with this predicate \emph{and} the subject that has already been mentioned in the caption. Introducing this indicator into the calculation of vocabulary scores reduces the probability of producing phrases that would require a specific type of fact being generated afterwards if no such facts are available in the knowledge context.

The three sets of scores are then concatenated and fed to a softmax layer, which produces an overall probability distribution over the tokens ($v_1\dots v_k$) $\in V$, ($e_1\dots e_n$) $\in G$ and ($f_1\dots f_m$) $\in K$ (see the diagram in Figure~\ref{fig:diagram}). The token with the highest probability is generated at position $t$.
\begin{equation}\label{eq:softmax_prob}
    \begin{split}
    &w_t = \argmax{w_i}P(w_i), w_i \in V \cup G \cup K \\
    &\text{ where }P(w_i) =\\
    &= \sigma_i(\text{ Concat}[y_{v_1}...y_{v_k}, y_{e_1}...y_{e_n}, y_{f_1}...y_{f_m}])\\
    \end{split}
\end{equation}

\section{Results and Discussion}
\label{sec:results}

\begin{table*}[ht]
    \centering
    \small
    \begin{tabular}{cccccccc}
        \hline
        & BLEU-1 & BLEU-2 & BLEU-3 & BLEU-4 & ROUGE & METEOR & CIDEr\\ \hline
        Standard \citep{mokady2021clipcap} & 0.33 & 0.09 & 0.02 & 0.0 & 5.62 & 2.45 & 0.14 \\
        No-knowledge & 16.29 & 8.43 & 4.38 & 2.25 & 23.14 & 7.24 & 1.75 \\
        Knowledge-from-image & 19.71 & 9.68 & 4.88 & 2.4 & 23.48 & 7.76 & 1.63 \\
        Knowledge-from-metadata & \textbf{26.84} & \textbf{17.98} & \textbf{12.85} & \textbf{9.5} & \textbf{33.47} & \textbf{13.52} & \textbf{32.41} \\ \hline
    \end{tabular}
    \caption{Metric scores of the different systems, measured on the test set (a single run).}
    \label{tab:metrics}
\end{table*}

We trained and tested our captioning system on the dataset described in Section~\ref{sec:dataset}. We evaluate its performance with standardly used automatic image captioning metrics: BLEU \citep{Papineni2002}, ROUGE \citep{Lin2004}, METEOR \citep{Denkowski2014} and CIDEr \citep{Vedantam2015}. We also present \emph{factual accuracy} as a separate focus of evaluation. Simply generating captions with verifiable statements is not enough --- they have to be accurate in their description of the entities in the image. We introduce a custom metric that measures the correctness of the generated captions against the DBpedia database.

\paragraph{Baselines}
For the first baseline, we examine the performance of a standard caption generation system \citep{mokady2021clipcap} on our test set. The standard system has no contextual components and was trained on the out-of-domain images from the MSCOCO dataset.

We create two more baseline captioning systems\footnote{The existing contextualized captioning systems (e.g.~the ones referenced in Section~\ref{sec:background}) are not suitable for direct use as baselines here. Most of these systems are specifically designed for certain types of data, which makes it unfeasible to run them on our dataset. For example, the news image captioning models require a news article text as a context for the image; in \citet{bai2021explain} the system is built and trained to generate specifically the descriptions of fine-art paintings.} and train them on our dataset.
The first one is a ``no-knowledge'' system, in which both geographic and knowledge components have been removed. Its performance represents the level that can be achieved by a standard image captioning pipeline with no additional contextualization.

The second baseline utilizes external knowledge based solely on the image. This system, which we call ``knowledge-from-image'', draws from the existing research on contextualized image captioning \citep{huang2020boost, Zhou2019}. Similarly to these works, our baseline uses a pre-trained object recognition model\footnote{We employ the VGG16 CNN model \citep{simonyan2014very} trained on the Places365 database \citep{zhou2017places} for scene recognition. The model's checkpoint is provided at \url{https://github.com/GKalliatakis/Keras-VGG16-places365}.} to detect objects and scenes in the image (e.g. ``beach'') and then queries the ConceptNet DB to retrieve the terms related to these objects (e.g. ``coast'', ``driftwood'', ``ocean''). We identify top 5 objects in every image and extract top 10 closest ConceptNet terms for each of them. We further pass their GloVe embeddings through a Transformer encoder and concatenate the result with the encoding of the image to construct a combined context representation for caption generation.

\subsection{Caption Quality Evaluation}

Table~\ref{tab:metrics} shows the comparison between the metric scores of our ``knowledge-from-metadata'' system and the three baselines. 
The standard system produces captions that are very different from the ground truth ones, which is reflected in the particularly low metric scores. This is expected, since the dataset it was trained on, as well as the architecture of the system itself, did not account for the context of the images and instead focused on the visual descriptions only. The no-knowledge and knowledge-from-image baselines achieve higher scores (the latter with a marginal improvement over the former in most of the metrics), but both are outperformed by our knowledge-from-metadata system\footnote{We also experimented with a ``geo-from-metadata'' baseline, which has access to the geographic context but not to the knowledge context. Its performance is higher than that of the other baselines, although still lower than our system's. For more details see Appendix~\ref{sec:appendix_geo_aware_baseline}.}.
All the improvements of the knowledge-from-metadata system are statistically significant (two-sample t-test, \textit{p} \textless 0.001). The difference in performance mainly stems from the ability of our system to produce contextually appropriate tokens from the external knowledge `vocabularies'; even though the same tokens are present in the vocabularies of the baselines and can be generated as regular vocabulary words, their generation in the baselines is dominated by noise and almost never matches the correct facts from the ground truth captions. The most radical improvements are in the CIDEr metric, which gives a higher weight to the words that are more informative according to the TF-IDF score; geographic names and fact-related tokens are usually rare in the corpus and highly informative, so they contribute a lot to this metric.

Table~\ref{tab:captions} shows examples\footnote{The images for these examples and additional examples from the test set are given in Appendix~\ref{sec:appendix_captions}.} of the captions generated by the knowledge-from-metadata system and the baselines, as well as the original human-written captions.
Here, the standard system from \citet{mokady2021clipcap} produces mostly accurate descriptions of what can be seen in the images but includes no references to their context. The no-knowledge and  knowledge-from-image baseline systems fail to generate correct geographic entity names and facts, as they draw all the caption tokens out of the general vocabulary. The captions generated by the knowledge-from-metadata system demonstrate that it is able to successfully use both geographic and knowledge contexts and produce accurate facts about relevant geographic entities.

\renewcommand{\arraystretch}{1.4}
\begin{table*}[ht]
    \centering
    \footnotesize
    \begin{tabularx}{\textwidth}{X}
    \hline
        \underline{Ground truth}: Kelso Bridge. Below the confluence of the Rivers Tweed and Teviot. John Rennie engineered the bridge, which was built between 1800 and 1803. \\
        \underline{Standard} \citep{mokady2021clipcap}: A river with a bridge and a train on it.\\
        \underline{No-knowledge}: the \textit{river dee}. \textit{farndon bridge} was opened in \textit{1339} by \textit{monks} from \textit{farndon bridge}.\\
        \underline{Knowledge-from-image}: \textit{chertsey bridge}. \textit{chertsey bridge} dates from \textit{1785}.\\
        \underline{Knowledge-from-metadata (ours)}: \textbf{kelso bridge}. the bridge over the \textbf{river tweed} was built in \textbf{1800}, and was designed by \textbf{john rennie the elder}.\\
        \hline
        \underline{Ground truth}: St. Mary's Lighthouse. On St Mary's Island, just north of Whitley Bay. Completed in 1898 the lighthouse remained operational until 1984 when it was superseded by modern navigational techniques. \\
        \underline{Standard} \citep{mokady2021clipcap}: A lighthouse with a white tower on top.\\
        \underline{No-knowledge}: the \textit{naze tower, walton on naze}. the \textit{naze tower} was built in \textit{1720} as a navigational aid to \textit{1720} as a navigational aid to aid \\
        \underline{Knowledge-from-image}: \textit{portland bill car park}. \textit{portland bill} is a popular destination for visitors to the lighthouse, built in \textit{1906}.\\
        \underline{Knowledge-from-metadata}: \textbf{st marys lighthouse}. the lighthouse was built in \textbf{1898} and is now a \textbf{grade ii listed} building. \\ 
        \hline
    \end{tabularx}
    \caption{Examples of the generated captions. Correct geographic references and facts are given in \textbf{bold}; incorrect ones are given in \textit{italics}.}
    \label{tab:captions}
\end{table*}
\renewcommand{\arraystretch}{1.0}
\begin{table}[ht]
    \centering
    \begin{tabular}{cccccccc}
        \hline
        & Accuracy, \% \\ \hline
        No-knowledge & 0.0 \\
        Knowledge-from-image & 0.0 \\
        Random fact & 53.35 \\
        Knowledge-from-metadata & \textbf{86.08} \\ \hline
    \end{tabular}
    \caption{Fact accuracy scores (a single run).}
    \label{tab:accuracy_scores}
\end{table}

\subsection{Generated Facts Accuracy}

An important indicator of the quality of the generated image captions is their factual accuracy. We measure it as the percentage of correct facts among all generated facts. To check the correctness of a given fact, we use a rule-based metric, which first verifies that the fact's subject entity is related to the image and then searches for common predicate-specific key phrases (e.g. ``designed by'') and fact objects from the knowledge context that should co-occur with the detected phrases (e.g. the name of the actual designer), with subsequent manual verification.

In addition to the no-knowledge and knowledge-from-image baselines introduced earlier, we also create a ``random fact'' baseline. It takes the captions generated by our system and replaces the fact-related tokens in each caption with the tokens randomly picked from the knowledge context, preserving the token type (e.g. replaces generated years with randomly picked years, generated names with randomly picked names, etc.). This creates quite a strong baseline because the probability of any entity from the knowledge context to be relevant to the image and thus to appear in the caption is high by design.

Table~\ref{tab:accuracy_scores} shows the accuracy scores of the three baselines and our knowledge-from-metadata system.
The no-knowledge and knowledge-from-image baselines did not produce any accurate facts. The strong random fact baseline's score is much higher but is still greatly outperformed by the knowledge-from-metadata system.

\section{Conclusions}

To mimic human behavior in describing an image, it is essential that image captioning systems take into account the context of the image and related real world knowledge.
In this paper, we have presented a novel method of contextualizing an image captioning pipeline with real world encyclopedic facts that are relevant to the image but are not directly inferable from it.
We compiled a new image captioning dataset with naturally produced informative captions and geographic metadata, which we use to obtain diverse image-specific knowledge from external data sources. The experiments on this dataset demonstrate the effectiveness of our approach: our captioning system is able to generate captions with correct facts about entities relevant to the image. Compared to a number of baseline systems, it achieves substantial improvements in the standard metrics as well as in the accuracy of the generated facts. In future work we plan to extend the proposed approach to other domains, such as medical image captioning \citep{ayesha2021automatic}, which would benefit greatly from incorporating external knowledge from medical databases.

\clearpage
\section*{Limitations}

In our approach, we rely on image location metadata to be the grounding ``anchor'' for retrieving external knowledge. Thus, if there is no image metadata or similar annotations available, our approach cannot be applied without substantial modifications. A side effect of this is that it is impossible to evaluate our system on the most common captioning datasets, such as MSCOCO or Flickr30k, which do not include image metadata.

Another limitation of our work is that we propose an encoding mechanism for external knowledge that is specific to the type and format of the knowledge utilized in our system ($<$subject, predicate, object$>$ triples). A different encoding strategy would have to be devised if the knowledge was supplied in a different format (e.g. free-form text from Wikipedia).

Finally, the performance of our system depends on the coverage of the external data sources. If there is no relevant knowledge in a database for a given image, the generated caption is unlikely to contain accurate encyclopedic facts.

\section*{Ethics Statement}
This paper presents a dataset of images, captions and geographic metadata collected from the Geograph project website. All the original data is licensed for reuse under the Creative Commons BY-SA 2.0 license. In compliance with the license terms, our dataset is distributed under the more recent version of the same license --- Creative Commons BY-SA 4.0. All the images in this paper are credited with a stamp that indicates the license, the author (the copyright owner) and the link to the original image.

This paper also presents an image captioning system, which can be deployed and run freely. The code of the system is made available under the Creative Commons BY-SA 4.0 license. We do not expect any specific harmful consequences of using the system or any potential for misuse.

\nocite{Anderson2018, Xu2015}
\bibliography{facts_paper}
\bibliographystyle{acl_natbib}

\appendix

\section{Dataset}
\label{sec:appendix_split}

\paragraph{Image selection} The main selection criterion for our image caption dataset is that each caption has to include at least one encyclopedic fact about a geographic entity relevant to the image. This ensures that the knowledge-aware captioning system has enough data to learn from. For practical reasons, we also select only captions that are no longer than 100 tokens.

\paragraph{Train/validation/test split} We split our captioning dataset into train, validation and test sets that constitute, respectively, 75\%, 12.5\% and 12.5\% of the whole dataset. In order to avoid assigning different photographs of the same geographic entities to both train and validation/test sets, we base the split on the latitude of the image location instead of splitting the dataset randomly. The photographs that were taken to the north of the 54.8975\textdegree{} latitude are assigned to the test set, between the 53.5706\textdegree{} and the 54.8975\textdegree{} latitude to the validation set, and the rest to the train set. With the latitude-based split, we ensure testing on previously unseen data, which helps to detect possible overfitting.

\paragraph{Caption pre-processing} During pre-processing, captions are converted to lower case and split into tokens with the standard NLTK Tokenizer package. Formatting artifacts are removed and some tokens are replaced with others for unification purposes (e.g. ``\&'' is replaced with ``and'', ``saint'' is replaced with ``st''). By simple string matching against the geographic and knowledge contexts, we identify geographic entities and facts in the captions and create a ``mask'' that shows the type of each token (an entity, a fact or a regular vocabulary word). The mask is supplied to the captioning system along with the pre-processed caption tokens and used for selecting the appropriate embedding function for each token (see Equation~\ref{eq:token_embed}).

\section{External Knowledge Statistics}
\label{sec:appendix_knowledge_stats}

\begin{table*}[ht]
    \centering
    \begin{tabular}{cccccccc}
        \hline
        & Per image & Per caption & Per entity \\ \hline
        Average number of facts & 36.49 & 2.2 & 7.34 \\
        Average number of predicates & 24.96 & 2.2 & 5.82 \\ \hline
    \end{tabular}
    \caption{Average numbers of facts/predicates.}
    \label{tab:fact_stats}
\end{table*}

Table~\ref{tab:fact_stats} provides the quantitative statistics of the external knowledge data used to train and evaluate our image captioning system. The average number per image shows how many distinct facts and predicates there are in the knowledge context for a given image. The average number per caption shows how many facts and predicates actually appear in the ground truth captions. Finally, the average number per entity shows how many unique facts and predicates a single subject (geographic entity) generally has in the knowledge context. The number of unique predicates is lower than the number of unique facts because in some cases several distinct facts may share a predicate, e.g. $<$Sadler's Wells Theatre, rebuilt, 1765$>$ and $<$Sadler's Wells Theatre, rebuilt, 1998$>$.

\begin{figure*}[ht]
    \centering
    \begin{center}
    \includegraphics[scale = 0.5]{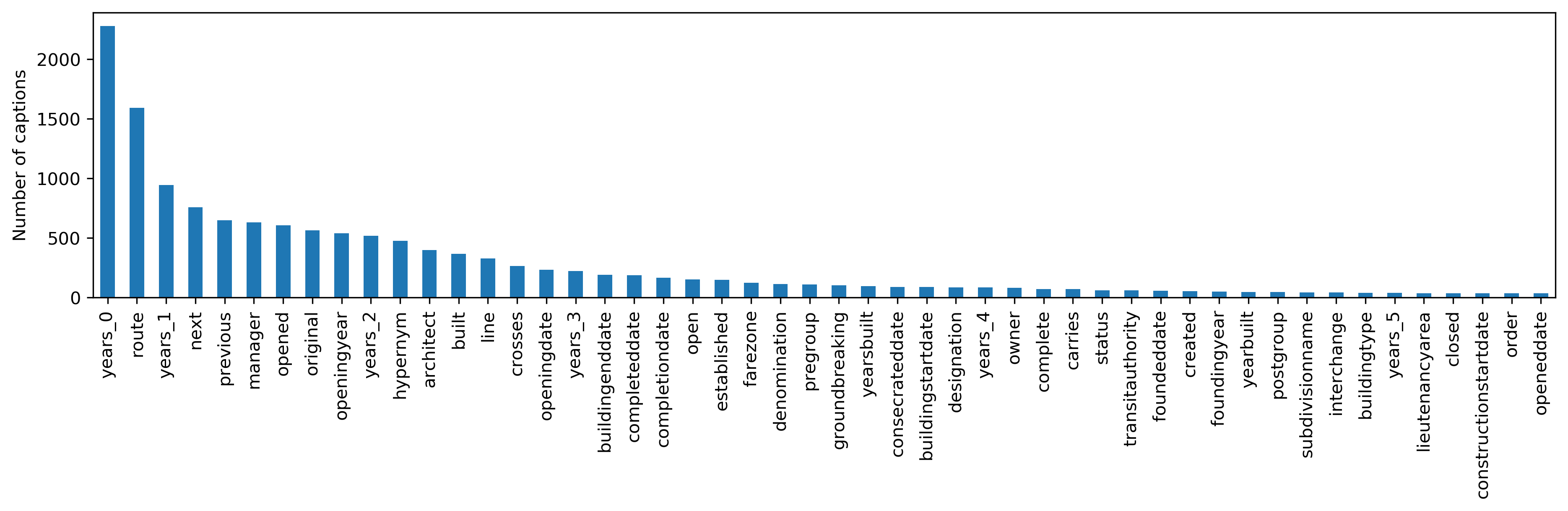}
    \caption{Distribution of the top 50 frequent original predicates.}
    \label{fig:pred_distribution_original}
    \end{center}
\end{figure*}

\begin{figure*}[ht]
    \centering
    \begin{center}
    \includegraphics[scale = 0.5]{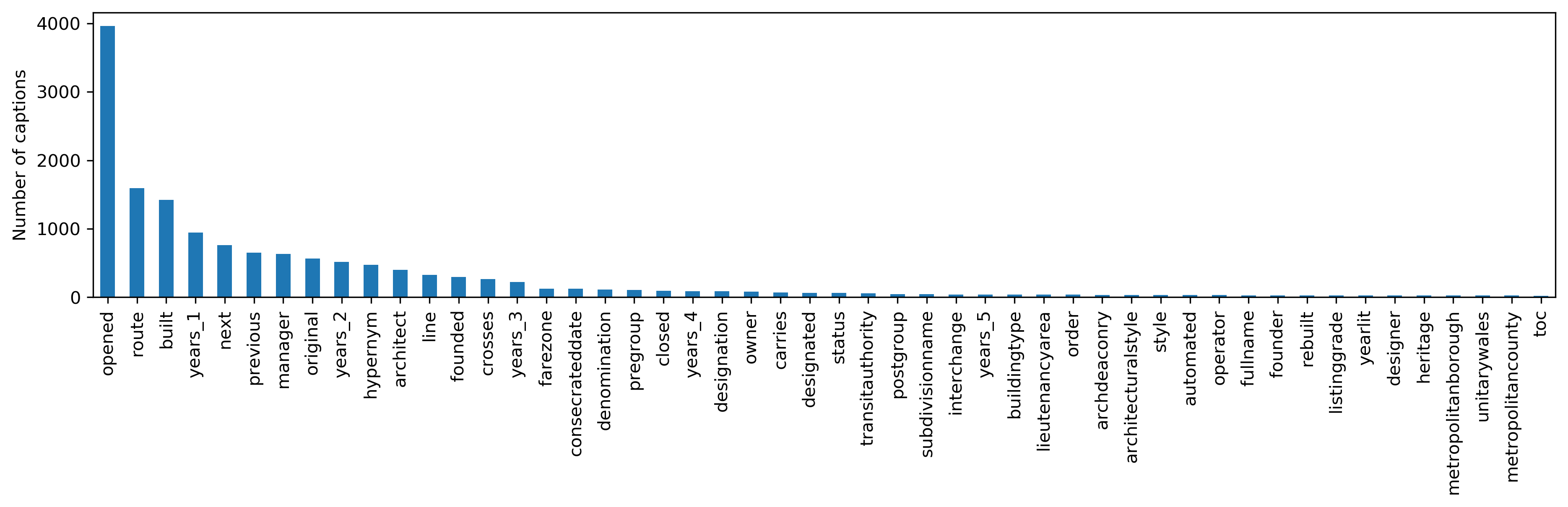}
    \caption{Distribution of the top 50 frequent predicates after merging the synonymous ones.}
    \label{fig:pred_distribution_merged}
    \end{center}
\end{figure*}

Figure~\ref{fig:pred_distribution_original} shows the distribution of the top 50 frequent predicates to appear in the captions. Many of the predicates displayed in the figure are in fact synonymous, i.e. reflect the same relation between the subject and object entities. For example, ``opened'' and ``openingyear'' are both used to link a geographic entity (the subject) and the year when it was opened (the object). Ideally, every unique relation between entities should be represented by a single predicate. On this account, we mapped frequent synonymous predicates to a single predicate that represents their common meaning, e.g. both ``opened'', ``openingyear'' are merged into ``opened''. The distribution of the top 50 most frequent predicates after merging the synonymous ones can be seen in Figure~\ref{fig:pred_distribution_merged}.

\section{Implementation}
\label{sec:appendix_implementation}

Our system is implemented in the PyTorch framework \citep{Paszke2019}, version 1.9.0, with CUDA version 11.0. The versions of the all the packages we use are provided in the requirements file that is distributed with the code. The system hyperparameters have been selected without automatic tuning; here we report the chosen values.

\paragraph{Encoder} All the images in the dataset are resized to 256x256 pixels to go through the pre-trained CNN image recognition module, which we do not fine-tune. The output of the image encoder has a size 14x14 with 2048 color channels.

The Transformer Encoder used for encoding the geographic and knowledge contexts is a standard PyTorch implementation, with 3 layers and 10 heads in a layer, with a dropout value of 0.5 and a feedforward network with a dimension size 512.

\paragraph{Context} The size of both the geographic embedding for the entities in the geographic context and the fact embedding for the facts in the knowledge context is 300, which matches the size of the pre-trained GloVe embeddings used for the vocabulary words. In the geographic embedding, the distance and the size parameters are used without any normalization. The azimuth of the angle between the object and the image location is normalized as follows:
\begin{equation}\label{eq:azimuth}
\begin{split}
 &a_i = \text{Concat}[a_{north};\:a_{east}] \\
 &\text{where}\\
 &a_{north} = |a_i| / 180 \\
 &a_{east} = 
\begin{cases}
   |90 - a_i| / 180, a_i >= -90 \\
   (90 + |a_i + 180|) / 180, \text{otherwise} \\
\end{cases}
\end{split}
\end{equation}

The normalization ensures that the azimuth values that are close to each other ``on the compass'' get similar values after normalization.

The types of the geographic entities (e.g. ``town'', ``river'') and the fact predicates (``architect'', ``opened'') are encoded with two separate embedding layers. Both are fine-tuned during the training of the image captioning system.

\paragraph{Decoder} The decoder follows a standard PyTorch implementation of a Transformer Decoder. Similarly to the encoder, the decoder has 3 layers and 10 heads in a layer, the dimension of the feedforward network is 512, the dropout value is 0.5. The embeddings of the vocabulary words are initialized from the GloVe embeddings of size 300, pre-trained on the Common Crawl data and fine-tuned during training.

\paragraph{Training} We train the system with the Adam optimizer with the learning rate of 4e-4. During backpropagation, the decoder gradients are clipped to the absolute value of 5.0. We use cross entropy loss; the early stopping is enabled after 20 consecutive epochs without a loss decrease. The training was carried out with a 4GB NVIDIA Quadro P1000 GPU, with an estimated total of 210 GPU hours for all the experiments, including baselines (a single training run takes approximately 3 hours, a single evaluation run --- approximately 3.5 minutes). The total number of trainable parameters in the network is 16,121,781.

\paragraph{Evaluation} The implementation of the standard captioning metrics (BLEU, ROUGE, METEOR, CIDEr) is adopted from \url{https://github.com/tylin/coco-caption}. For one of the baselines we use a pre-trained ClipCap captioning model, with the implementation and checkpoint provided at \url{https://github.com/rmokady/CLIP_prefix_caption}.

\section{Alternative Architectures}
\label{sec:appendix_alt}

\paragraph{No predicate indicator} In an alternative setup without the predicate indicator $p_{ind}$, the scores calculation from Equation~\ref{eq:scores} is modified for the vocabulary words $v_1...v_k \in V$ as follows:
\begin{equation}\label{eq:scores_without_p_ind}
    y_{v_1}...y_{v_k} = h_t \; W_{vocab}
\end{equation}

In this case, the influence of the knowledge context on vocabulary word selection is not as strong. This leads to a decrease in the generated fact accuracy, which is measured at 80.74\% (as compared to the fact accuracy of 86.08\% when the predicate indicator is used). A typical mistake with this setup is generating a phrase from the vocabulary that warrants a certain type of fact that is not present in the knowledge context. For example, generating ``elie house, founded in 1697'' when actually Elie House was \textit{built} in 1697. Since there is no fact in the corresponding knowledge context with the subject ``Elie House'' and the predicate ``founded'', generating ``founded in'' after ``elie house'' would very likely lead to producing an incorrect fact. Accounting for the knowledge context while generating vocabulary words is the purpose of the predicate indicator.  

\paragraph{No geo entity indicator} Without the geo entity indicator $g_{ind}$, the scores calculation from Equation~\ref{eq:scores} is modified for the facts $f_1...f_m \in K$ as follows:
\begin{equation}\label{eq:scores_without_g_ind}
    y_{f_1}...y_{f_m} = (EmbK \; \textsc{diag}(h_t))\: \vec{w}_{f}
\end{equation}

Without this indicator, the model is more prone to generating facts, the subjects of which have not been mentioned in the caption. This often results in a mistake when the model selects a fact with an appropriate predicate but an incorrect subject, e.g. generating ``neidpath viaduct, built in 1263'' with the following two facts in the knowledge context: $<$Neidpath Viaduct, built, 1863$>$ and $<$Neidpath Castle, built, 1263$>$. The architecture without the geo entity indicator scores 76.86\% in the generated fact accuracy, almost 10 percentage points lower than when the geo entity indicator is present.

\section{Geographic-Knowledge-from-Metadata Baseline}
\label{sec:appendix_geo_aware_baseline}

In order to evaluate the significance of the knowledge context component in our image captioning system, we conduct an experiment with another baseline, called ``geo-from-metadata''. This baseline has access to the geographic context but not to the knowledge context: the output of the encoder is the concatenation of the image representation and the geographic context encoding alone, and during caption generation the decoder can only pick from vocabulary tokens and geographic entity names. The difference between the performance level of the geo-from-metadata and knowledge-from-metadata systems demonstrates the impact the knowledge context has on the generated captions.

The standard captioning metrics scores of the geo-from-metadata system\footnote{BLEU-1 --- 21.91, BLEU-2 --- 13.25, BLEU-3 --- 8.35, BLEU-4 --- 5.32, ROUGE --- 30.19, METEOR --- 10.98, CIDEr --- 16.59.} show a significant improvement over the no-knowledge baseline, which does not include any contextualization at all; however, the scores remain much lower than the ones achieved by our knowledge-from-metadata system. Examples of the captions generated by the geo-from-metadata system are given in Appendix~\ref{sec:appendix_captions}. As is evident from the captions, the geo-from-metadata system can utilize the geographic context available to it to produce accurate geographic references but, not being able to use the knowledge context, does not produce correct encyclopedic facts. With a few coincidental right guesses, the fact accuracy on a test set is measured at 6.56\%. Thus, incorporating knowledge context is shown to be crucial for generating captions with accurate encyclopedic facts about entities relevant to the image.

\section{Examples of the Generated Captions}
\label{sec:appendix_captions}

\renewcommand{\arraystretch}{1.5}
\begin{table*}[ht]
    \centering
    \footnotesize
    \begin{tabularx}{\textwidth}{lcX}
    \hline
        \multirow{4}{*}{(a)} & \multirow{4}{*}{
        \raisebox{-0.9\totalheight}[0\height][\height]
        {\includegraphics[width=2.5cm]{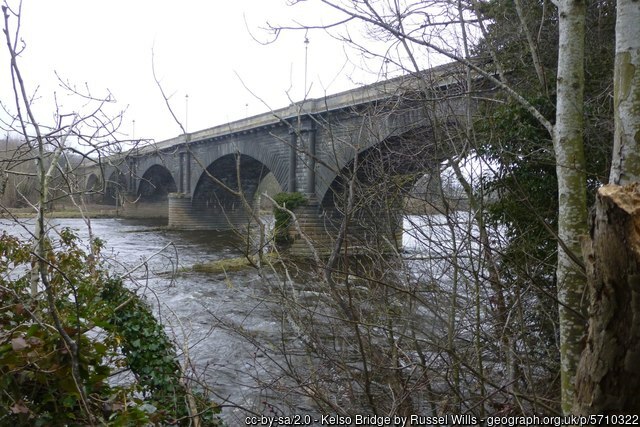}}
        } & \underline{Ground truth}: Kelso Bridge. Below the confluence of the Rivers Tweed and Teviot. John Rennie engineered the bridge, which was built between 1800 and 1803. \\
        & & \underline{Standard} \citep{mokady2021clipcap}: A river with a bridge and a train on it.\\
        & & \underline{No-knowledge}: the \textit{river dee}. \textit{farndon bridge} was opened in \textit{1339} by \textit{monks} from \textit{farndon bridge}.\\
        & & \underline{Knowledge-from-image}: \textit{chertsey bridge}. \textit{chertsey bridge} dates from \textit{1785}.\\
        & & \underline{Geo-from-metadata}: the \textbf{river tweed}. the bridge is a \textit{grade ii listed} building. the bridge was built in \textit{1826} and completed in \textit{1805}.\\
        & & \underline{Knowledge-from-metadata}: \textbf{kelso bridge}. the bridge over the \textbf{river tweed} was built in \textbf{1800}, and was designed by \textbf{john rennie the elder}.\\
        \hline
        \multirow{4}{*}{(b)} & \multirow{4}{*}{
        \raisebox{-0.9\totalheight}[0\height][\height]
        {\includegraphics[width=2.5cm]{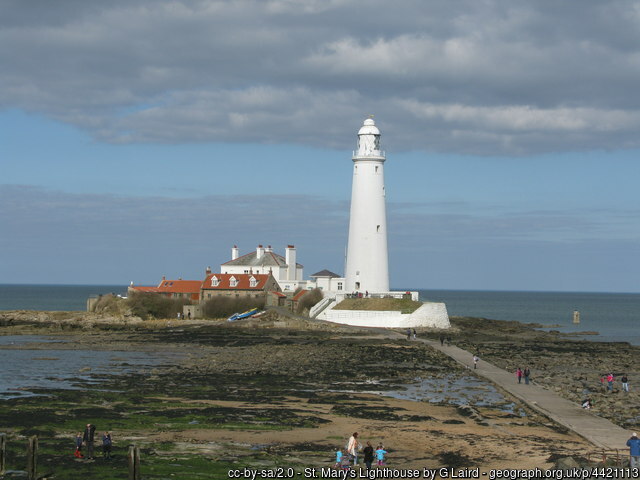}}
        } & \underline{Ground truth}: St. Mary's Lighthouse. On St Mary's Island, just north of Whitley Bay. Completed in 1898 the lighthouse remained operational until 1984 when it was superseded by modern navigational techniques.\\
        & & \underline{Standard} \citep{mokady2021clipcap}: A lighthouse with a white tower on top.\\
        & & \underline{No-knowledge}: the \textit{naze tower, walton on naze}. the \textit{naze tower} was built in \textit{1720} as a navigational aid to \textit{1720} as a navigational aid to aid \\
        & & \underline{Knowledge-from-image}: \textit{portland bill car park}. \textit{portland bill} is a popular destination for visitors to the lighthouse, built in \textit{1906}.\\
        & & \underline{Geo-from-metadata}: \textbf{st marys lighthouse}. the lighthouse was built in \textit{1833}, and is \textbf{grade ii listed}.\\
        & & \underline{Knowledge-from-metadata}: \textbf{st marys lighthouse}. the lighthouse was built in \textbf{1898} and is now a \textbf{grade ii listed} building.\\ 
        \hline
        \multirow{4}{*}{(c)} & \multirow{4}{*}{
        \raisebox{-0.9\totalheight}[0\height][\height]
        {\includegraphics[width=2.5cm]{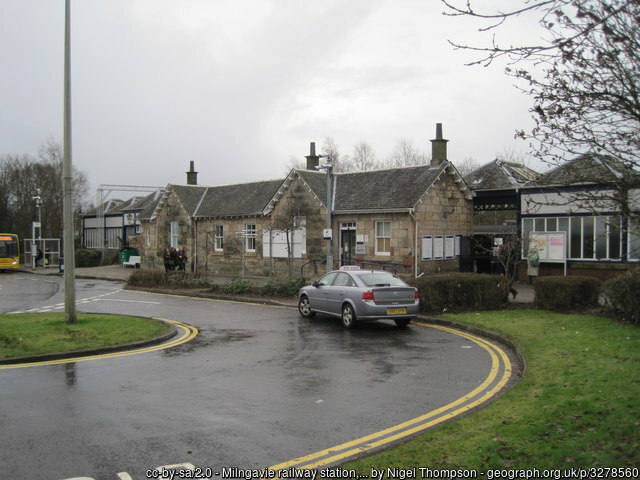}}
        } & \underline{Ground truth}: Milngavie railway station, East Dunbartonshire. Opened in 1863 by the Glasgow \& Milngavie Junction Railway. View south west at forecourt. \\
        & & \underline{Standard} \citep{mokady2021clipcap}: A car driving down a street past a building.\\
        & & \underline{No-knowledge}: \textit{st johns wood church, threapwood}. the \textit{church} was built in \textit{1848} by the architect \textit{john soane}. 
        \textit{grade ii} listed building (english heritage building id: link).\\
        & & \underline{Knowledge-from-image}: \textit{staines station}. \textit{staines station} is a station on the \textit{waterloo to reading line, and the junction of windsor}. the station was opened in \textit{1848}.\\
        & & \underline{Geo-from-metadata}: \textbf{milngavie station}. the station is located on the \textbf{milngavie branch}. the station is located on the site of the station, and the station was opened in \textbf{1863}.\\
        & & \underline{Knowledge-from-metadata}: \textbf{milngavie station}. the station was opened in \textbf{1863}. it is now part of the \textbf{abellio scotrail}.\\
        \hline
        \multirow{4}{*}{(d)} & \multirow{4}{*}{
        \raisebox{-0.99\totalheight}[0\height][\height]
        {\includegraphics[width=2.1cm]{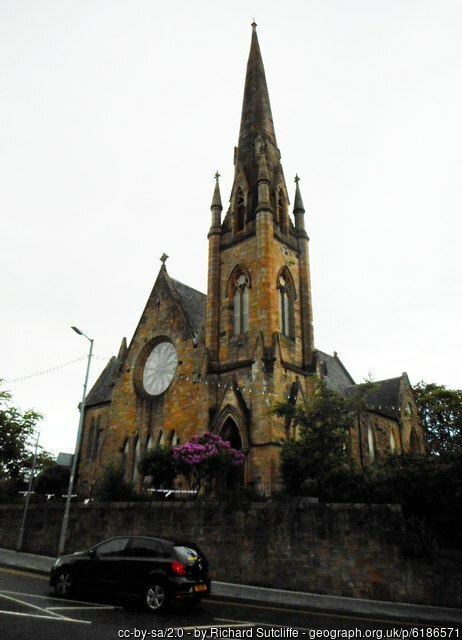}}
        } & \underline{Ground truth}: Lenzie Old Parish Church. A Category C listed church [Link], dating from 1874. \\
        & & \underline{Standard} \citep{mokady2021clipcap}: A church with a clock tower and a cross on top.\\
        & & \underline{No-knowledge}:  \textit{st peters church, colwyn bay}. \textit{st asaph}.  \textit{st pauls church} was consecrated in \textit{1887}.\\
        & & \underline{Knowledge-from-image}: \textit{christ church, heaton norris}. \textit{christ church} was built in \textit{1846}. a developed \textit{early english style}. the church was built in \textit{1846}. the church was designed by \textit{sir christopher wren}.\\
        & & \underline{Geo-from-metadata}: \textbf{lenzie old parish, lenzie}. built in \textit{1846}, the church was built in \textit{1846}.\\
        & & \underline{Knowledge-from-metadata}: \textbf{lenzie old parish church , lenzie}. \textbf{lenzie old parish church} was built in \textbf{1874} and designed by \textbf{clarke and bell}. it is now a \textit{grade ii} listed.\\
        \hline
    \end{tabularx}
    \caption{Examples of the generated captions. Correct geographic references and facts are given in \textbf{bold}; incorrect ones are given in \textit{italics}.
    \newline {\scriptsize Image references: \newline(a) \url{https://www.geograph.org.uk/photo/5710322}; \newline(b) \url{https://www.geograph.org.uk/photo/4421113}; \newline(c) \url{https://www.geograph.org.uk/photo/3278560}; \newline(d) \url{https://www.geograph.org.uk/photo/6186571}}} 
    \label{tab:captions_appendix}
\end{table*}
\renewcommand{\arraystretch}{1.0}

\end{document}